%% file: acl2023.tex
\pdfoutput=1
\documentclass[11pt]{article}

\usepackage{ACL2023}
\usepackage{times}
\usepackage{latexsym}
\usepackage{url}
\usepackage{arydshln}
\usepackage{enumitem}
\usepackage{float}
\input{math_commands.tex}

\input{macro}

\usepackage{amssymb}
\usepackage{amsmath}
\usepackage{tikz,pgfplots}
\usepackage{algorithm}
\usepackage{algorithmic}
\renewcommand{\algorithmiccomment}[1]{\bgroup\hfill//~#1\egroup}
\usepackage[switch]{lineno}  %
\usepackage{caption}
\usepackage{subcaption}
\usepackage{booktabs, multirow}
\usepackage{comment}
\usepackage{balance}
\usepackage{cleveref}
\crefname{section}{§}{§§}
\Crefname{section}{§}{§§}

\usepackage[T1]{fontenc}
\usepackage[utf8]{inputenc}

\usepackage{microtype}

\usepackage{inconsolata}
\usepackage{pgfplots}
\usepackage{tikz}
\usepackage{mathdots}
\pgfplotsset{compat=1.14}
\definecolor{bblue}{HTML}{4F81BD}
\definecolor{rred}{HTML}{C0504D}
\definecolor{ggreen}{HTML}{9BBB59}
\definecolor{ppurple}{HTML}{9F4C7C}
\pgfplotsset{
  compat=1.14,
  legend entry/.initial=,
  every axis plot post/.code={%
      \pgfkeysgetvalue{/pgfplots/legend entry}\tempValue
      \ifx\tempValue\empty
          \pgfkeysalso{/pgfplots/forget plot}%
      \else
          \expandafter\addlegendentry\expandafter{\tempValue}%
      \fi
  },
}
\title{A Two-Stage Decoder for Efficient ICD Coding}

\author{Thanh-Tung Nguyen$^{\dagger}$, Viktor Schlegel$^{\dagger}$, Abhinav Kashyap$^{\dagger}$, Stefan Winkler$^{\dagger}$$^\P$\\
  $^\dagger$ASUS Intelligent Cloud Services (AICS), Singapore \\
  $^\P$Department of Computer Science, National University of Singapore \\
  \texttt{\{thomas\_nguyen;viktor\_schlegel;abhinav\_kashyap;stefan\_winkler\}@asus.com}
}

\begin{document}
\maketitle
\begin{abstract}
Clinical notes in healthcare facilities are tagged with the International Classification of Diseases (ICD) code; a list of classification codes for medical diagnoses and procedures. ICD coding is a challenging multilabel text classification problem due to noisy clinical document inputs and long-tailed label distribution. Recent automated ICD coding efforts improve performance by encoding medical notes and codes with additional data and knowledge bases. However, most of them do not reflect how human coders generate the code: first, the coders select general code categories and then look for specific subcategories that are relevant to a patient's condition. Inspired by this, we propose a two-stage decoding mechanism to predict ICD codes. Our model uses the hierarchical properties of the codes to split the prediction into two steps: At first, we predict the parent code and then predict the child code based on the previous prediction. Experiments on the public MIMIC-III data set show that our model performs well in single-model settings without external data or knowledge.
\end{abstract}
% \vspace{-1.5em}
\section{Introduction}
Medical records and clinical documentation contain critical information about patient care, disease progression, and medical operations. After a patient's visit, medical coders process them and extract key diagnoses and procedures according to the International Classification of Diseases (ICD) system \citep{ICD}. Such codes are used for predictive modeling of patient care and health status, for insurance claims, billing mechanisms, and other hospital operations \citep{Tsui2002ValueOI}.

Although the healthcare industry has seen many innovations, many challenges related to manual operations still remain. One of these challenges is manual ICD coding, which requires understanding long and complex medical records with a vast vocabulary and sparse content. Coders must select a small subset from a continuously expanding set of ICD codes (from around 15,000 codes in ICD 9 to around 140,000 codes in ICD 10 \cite{ICD102016}). Therefore, manual ICD coding may result in errors and cause revenue loss or improper allocation of care-related resources. Thus, automated ICD coding has received attention not only from the industry but also from the academic community.

\begin{figure*}[t!]
\centering
\includegraphics[width=0.8\textwidth]{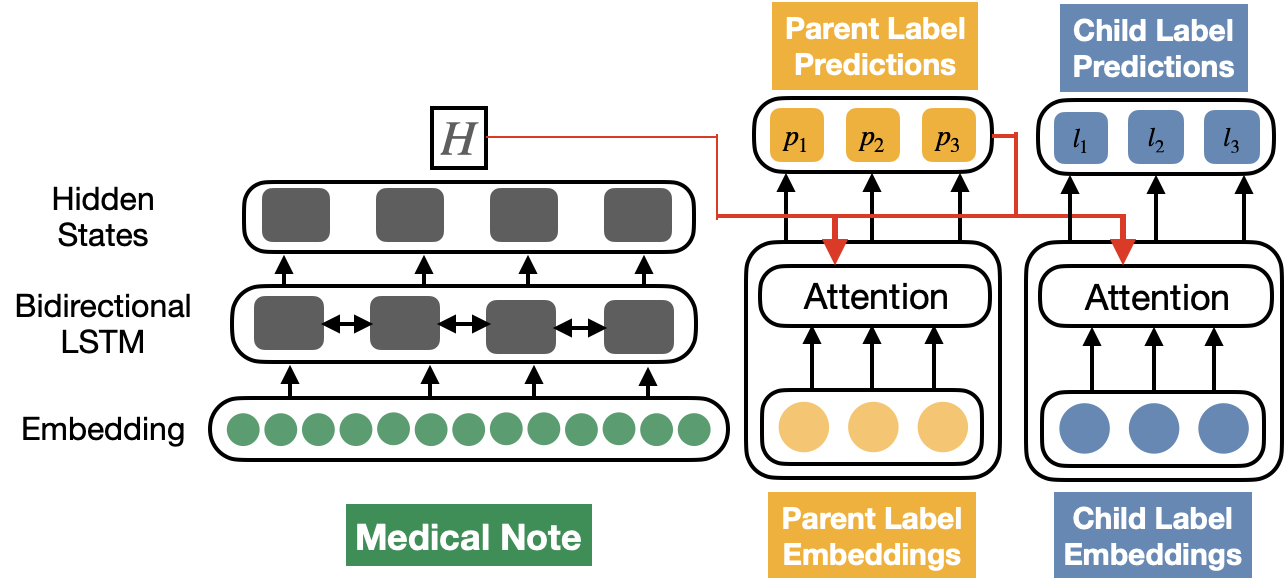}
% \vspace{-1em}
\caption{\small The architecture of our proposed two-stage decoding model. \textbf{Encoder (left side)}: the Medical note is embedded and passed through a bidirectional LSTM. \textbf{Decoder (right side)}: The first level parent label decoding uses parent label embeddings and attention over the note. The second-level child label predictions are made by attending to the parent label and the note. }
% \red{make it single column. we may need more space}}
\label{fig:model architecture}
\end{figure*}

Before the rise of deep learning methods, automated ICD coding methods applied rules or decision tree-based methods \citep{rulebaseicd2008,scheurwegs2017selecting}. The focus has now changed to neural networks using two strands of approaches. The first encodes medical documents using pretrained language models \citep{li2020multirescnn, liu-etal-2021-effective}, adapts pretrained language models to make them suitable for the clinical domain \citep{lewis-etal-2020-pretrained} or injects language models with medical knowledge such as taxonomy, synonyms, and abbreviations of medical diseases \citep{Yang2022KnowledgeIP,yuan-etal-2022-code}. The second improves the representation of pretrained language models, by capturing the relevance between the document and the label metadata such as their descriptions \citep{mullenbach-etal-2018-explainable,ijcai2020-461-vu,pmlr-v149-kim21a, zhou-etal-2021-automatic}, co-occurrences \citep{cao-etal-2020-hypercore}, hierarchy \citep{falis-etal-2019-ontological,ijcai2020-461-vu, LIU2022104161}, or thesaurus knowledge, such as synonyms \citep{yuan-etal-2022-code}. 
Although these approaches are supposed to alleviate problems specific to medical coding such as special vocabulary, a large set of labels, etc., they fall short.

Intuitively, human coders generate the code in two stages: first, the coders select the general codes and then look for specific subcategories that are relevant to a patient's condition. The advantage of adapting this approach to neural networks is that at each stage of the prediction, we deal with a smaller output space and we can have more confidence when predicting the next stage.
% While trivial and simple, methods of this type still deal with a large number of output spaces and face long-tail problems.

% The neural network methods used for multilabel classification focus on two main techniques: encoding documents effectively --using pretrained language models trained on clinical data --; use information from labels such as their description, co-occurrences with other labels, and hierarchy to enrich the representation of documents. Although label embedding is supposed to alleviate the problem of a large number of labels with few of them having enough examples, it falls short. Capturing the interaction between the label meta-data and the documents using attention is non trivial and cumbersome.

Therefore, in this paper, we introduce a simple two-stage decoding framework \footnote{https://github.com/thomasnguyen92/two-stage-decoder-icd} for ICD coding to mimic the processes of human coders. Our approach leverages the hierarchical structure of the ICD codes to decode, i.e., having parent-child relationships. The first stage predicts the parent codes;  the second stage uses the document representation and the predicted parent codes to predict the child codes. Experiments with MIMIC-III data sets demonstrate the effectiveness of our proposed method. In particular, our simple method outperforms models that use external knowledge and data. Since our models (\Cref{fig:model architecture}) are based on LSTMs, they require less computing power and can be trained faster %a shorter training time 
than other larger models.

\section{Two-stage Decoding Framework}
%\vspace{-0.5em}
ICD codes follow a hierarchical structure. In this work, we consider characters before the dot (.) in the ICD code as the parent label and the code that has to be predicted as the child label. For example, for the child label \textbf{39.10} about \textbf{Actinomycosis of lung}, its parent code is \textbf{39} representing \textbf{Actinomycotic infections}. Let $\sP$ and $\sL$ represent the sets of parent nodes and child codes for a medical note $\vx$, respectively. It is worth noting that if we know the child codes, we can use the above definition to find the corresponding parent codes. This means that knowing $\sL$ is equivalent to knowing both $\sL$ and $\sP$. Then the probability of the child labels is:
% \vspace{-0.5em}
\begin{align}
\begin{split}
P_{\theta}(\sL|\vx) & \sim  P_{\theta}(\sL,\sP|\vx) \\
   & =  P_{\theta}(\sL|\sP,\vx) P_{\theta}(\sP|\vx) 
\end{split}
\label{eq1}
\end{align}
\normalsize

This factorization allows us to compute the prediction scores of the parent codes first, and then, conditioned on them and the document, we can obtain the prediction score of the child codes. Therefore, we can %also 
model the ICD coding task using a decoder framework where we generate parent labels before predicting child labels. In this case, we adapt the decoder framework to the multilabel problem setting, where at each decoding stage, we predict multiple labels at once, instead of one label at a time like a standard decoder.

% There are two closely related work but different:
% Joint LAAT (https://arxiv.org/pdf/2007.06351.pdf does not actually consider the parent representations in making the child label prediction.

% https://aclanthology.org/D19-6220.pdf, https://arxiv.org/pdf/2208.02301.pdf also considers even grandparent level, however, the way they do it is not decoding stage by stage but just finding the representation of the child nodes.
%\vspace{-0.5em}
\subsection{Model Architecture}
%\vspace{-0.5em}
We now describe the components of our parsing model: the document encoder, the first decoding stage for the parent code, and the second decoding stage for the child code.

% \begin{table*}[!t]
% \centering
% \resizebox{1.1\columnwidth}{!}{%
% \begin{tabular}{lrrrrrrr}\toprule
% \multirow{2}{*}{Model} &\multicolumn{2}{c}{AUC} &\multicolumn{2}{c}{F1} &Precision \\\cmidrule{2-6}
% &Macro &Micro &Macro &Micro &P@8 & \\\midrule
% \multicolumn{6}{c}{\textbf{Single models}} \vspace{0.2em}\\
% MultiResCNN\citep{li2020multirescnn} &91.0 &98.6 &9.0 &55.2 &73.4  \\
% MSATT-KG \citep{10.1145/3357384.3357897}&91.0 &98.6 &8.5 &55.3 &72.8  \\
% JointLAAT \citep{ijcai2020-461-vu}&92.1 &98.8 &10.2 &57.5 &73.5  \\
% Our Model &94.6 &99.0 &10.5 &58.4 &74.4  \\
% \hline
% \multicolumn{6}{c}{\textbf{Models with External Data/Knowledge}} \vspace{0.2em}\\
% MSMN \citep{yuan-etal-2022-code}&95.0 &99.2 &10.3 &58.2 &74.9  \\
% PLM-ICD &92.6 &98.9 &10.4 &59.8 &77.1  \\
% KEPTLongformer \citep{Yang2022KnowledgeIP}&- &- &11.8 &59.9& 77.1\\
% \bottomrule
% \end{tabular}
% }
% \caption{Results on the MIMIC-III-Full test set.}\label{tab:result_full}
% \end{table*}

\begin{table*}[!t]
\centering
\resizebox{2\columnwidth}{!}{%
\begin{tabular}{lrrrrrrrrrrrr}\toprule
& \multicolumn{6}{c}{\bfseries MIMIC-III-Full} &\multicolumn{6}{c}{\bfseries MIMIC-III-50} \\
\multirow{2}{*}{Model} &\multicolumn{2}{c}{AUC} &\multicolumn{2}{c}{F1} & Precision & &\multicolumn{2}{c}{AUC} &\multicolumn{2}{c}{F1} &Precision  \\\cmidrule{2-6}\cmidrule{8-12}
&Macro &Micro &Macro &Micro &P@8 & &Macro &Micro &Macro &Micro &P@5 & \\\midrule
%\multicolumn{6}{c}
{\textbf{Single models}} \vspace{0.2em}\\
MultiResCNN \citep{li2020multirescnn} &91.0 &98.6 &9.0 &55.2 &73.4 & &89.30 &92.04 &59.29 &66.24 &61.56 \\
MSATT-KG \citep{10.1145/3357384.3357897}&91.0 &98.6 &8.5 &55.3 &72.8 & &91.40 &93.60 &63.80 &68.40 &64.40  \\
JointLAAT \citep{ijcai2020-461-vu}&92.1 &98.8 &10.2 &57.5 &73.5 & &92.36 &94.24 &66.95 &70.84 &66.36 \\
Our Model &94.6 &99.0 &10.5 &58.4 &74.4 & &92.58 &94.52 &\textbf{68.93} &71.83 &66.72 \\
\midrule
%\multicolumn{12}{c}
{\textbf{Models with External Data/Knowledge}} \vspace{0.2em}\\
MSMN \citep{yuan-etal-2022-code}&95.0 &99.2 &10.3 &58.2 &74.9 & & 92.50 &94.39 &67.64 &71.78 &67.23  \\
PLM-ICD \citep{huang-etal-2022-plm} &92.6 &98.9 &10.4 &59.8 &77.1 & &90.23 &92.44 &65.23 &69.26& 64.61 \\
KEPTLongformer \citep{Yang2022KnowledgeIP}&- &- &11.8 &59.9& 77.1 & &92.63 &94.76 &68.91 &72.85& 67.26\\
\bottomrule
\end{tabular}
}
\caption{Results on the MIMIC-III-Full and MIMIC-III-50 test sets. All our experiments are run five different random seeds and we report the mean results. The results of other models, except PLM-ICD, are collected from \citet{Yang2022KnowledgeIP}. For PLM-ICD, we follow the authors' instructions to reproduce the results.}\label{tab:result_full}
\end{table*}

% \begin{table*}[t!]
% \centering
% \resizebox{1.1\columnwidth}{!}{%
% \begin{tabular}{lrrrrr}\toprule
% \multirow{2}{*}{Model} &\multicolumn{2}{c}{AUC} &\multicolumn{2}{c}{F1} &Precision  \\\cmidrule{2-6}
% &Macro &Micro &Macro &Micro &P@5 \\\midrule
% \multicolumn{5}{c}{\textbf{Single models}} \vspace{0.2em}\\
% MultiResCNN \citep{li2020multirescnn} &89.30 &92.04 &59.29 &66.24 &61.56 \\
% MSATT-KG \citep{10.1145/3357384.3357897}&91.40 &93.60 &63.80 &68.40 &64.40 \\
% JointLAAT \citep{ijcai2020-461-vu}&92.36 &94.24 &66.95 &70.84 &66.36 \\
% % Effective CAN &92.0 &94.5 &66.8 &71.7 &66.4 & - \\
% % Our Model &92.57 &94.44 &\textbf{69.01} &71.83 &66.47 &13 \\
% Our Model &92.58 &94.52 &\textbf{68.93} &71.83 &66.72 \\
% \hline
% \multicolumn{5}{c}{\textbf{Models with External Data/Knowledge}} \vspace{0.2em}\\
% MSMN \citep{yuan-etal-2022-code} & 92.50 &94.39 &67.64 &71.78 &67.23\\
% KEPTLongformer \citep{Yang2022KnowledgeIP} &92.63 &94.76 &68.91 &72.85& 67.26\\
% \bottomrule
% \end{tabular}
% }
% \caption{Results on the MIMIC-III-50 test set. The results are the mean over five runs with different random seeds.}\label{tab:result_common_50}
% \end{table*}

% \vspace{-0.5em}
\paragraph{Document Encoder}
Given a medical note of $n$ tokens $\vx = (x_1, \ldots, x_n)$, we embed each token in the document in a dense vector representation. Subsequently, the token representations are passed to a single-layer BI-LSTM encoder to obtain the contextual representations $[h_1, h_2, \ldots, h_n]$. Finally, we obtain the encoding matrix $\mH \in \sR^{n \times d}$.

\vspace{-0.5em}
\paragraph{First Decoding Stage}
At this stage, similar to \citet{ijcai2020-461-vu}, we take the embedding of all parent labels $\mP \in \sR^{|L_P| \times d_e}$ to compute the attention scores and obtain the label-specific representations as:
\begin{align*}
\begin{split}
& \text{s}(\mP,\mH) = \mP \text{ tanh}(\mW\mH^T) \\
& \text{att}(\mP, \mH) = \softmax(\text{s}(\mP,\mH))\mH \\
& P(\sP|\vx) = \sigmoid(\text{rds} (\mV \boldsymbol{\odot} \text{att}(\mP, \mH))) \\
\end{split}
\end{align*}
\noindent where $\softmax(\cdot)$, $\sigmoid(\cdot)$, $\text{rds}(\cdot)$ denote row-wise \textit{softmax}, \textit{sigmoid}, \textit{reduce sum} in last dimension  operations% over each row of the matrix $\mA$
;~$\mW \in \sR^{d_e \times d}$ are the weight parameters to perform linear transformations and $\mV \in \sR^{|L_P| \times d}$ is the weight matrix of a label-wise fully connected layer which yields the parent label logits where $\odot$ is element-wise product. %$\text{s}(\mP,\mH)$.
% \vspace{-0.5em}
\paragraph{Second Decoding Stage}
At this stage, we take the label embeddings of all child labels $\mL~\in~\R^{|L| \times d_e}$ and the probabilities of predicted parent labels from the previous stage as input, and obtain the label-specific representations as per:

\begin{align*}
\begin{split}
\text{s}(\mL,\mP) &= \mL \text{ tanh}(\mW_P \boldsymbol{\odot}(P(\sP|\vx))^T)\\
\text{att}(\mL, \mP) &= \softmax(\text{s}(\mL,\mP))\mP \\
\text{s}(\mL,\mH) &= \mL \text{ tanh}(\mW_L\mH^T) \\
\text{att}(\mL, \mH) &= \softmax(\text{s}(\mL,\mH))\mH \\
P(\sL|\sP,\vx) &= \sigmoid(\text{rds}(\mV_{LH} \boldsymbol{\odot} \text{att}(\mL, \mH)\\
&+\mV_{LP} \boldsymbol{\odot} \text{att}(\mL, \mP)))
\end{split}
\end{align*}
%\vspace{-0.25em}
\noindent where we perform a `soft' embedding of the parent labels by taking the element-wise product between matrix $\mW_P \in \sR^{d_e\times |L_P|}$ with the sigmoid probabilities of parent labels. $\mV_{LH},\mV_{LP}  \in \sR^{|L_P| \times d}$ are the weight matrices of two label-wise fully connected layers that compute the child label logits.
% The resulting attention representation and the label representations are transformed using fully connected layers ($\mV_L,\mV_{LP}  \in \sR^{|L_P| \times d}$) and summed up to get the final logits of the child labels. %$\text{s}(\mL,\mP)$.
% \vspace{-0.5em}
\paragraph{Training Objective \& Inference}
The total training loss is the sum of the binary cross-entropy losses to predict the parent and child labels:
\begin{align}
\Ls_{\text{total}} (\theta) = \Ls_{P} (\theta) + \Ls_{L} (\theta)
\end{align}
\normalsize

For inference, we assign a child label to a document if the corresponding parent label score and the child label score are greater than predefined thresholds. 
%\vspace{-0.5em}
\section{Experiments}
\label{sec:experiments}

% \begin{table*}[t!]
% \centering
% \resizebox{1.1\columnwidth}{!}{%
% \begin{tabular}{lrrrrr}\toprule
% \multirow{2}{*}{Model} &\multicolumn{2}{c}{AUC} &\multicolumn{2}{c}{F1} &Precision  \\\cmidrule{2-6}
% &Macro &Micro &Macro &Micro &P@5 \\\midrule
% \multicolumn{5}{c}{\textbf{Single models}} \vspace{0.2em}\\
% MultiResCNN \citep{li2020multirescnn} &89.30 &92.04 &59.29 &66.24 &61.56 \\
% MSATT-KG \citep{10.1145/3357384.3357897}&91.40 &93.60 &63.80 &68.40 &64.40 \\
% JointLAAT \citep{ijcai2020-461-vu}&92.36 &94.24 &66.95 &70.84 &66.36 \\
% % Effective CAN &92.0 &94.5 &66.8 &71.7 &66.4 & - \\
% % Our Model &92.57 &94.44 &\textbf{69.01} &71.83 &66.47 &13 \\
% Our Model &92.58 &94.52 &\textbf{68.93} &71.83 &66.72 \\
% \hline
% \multicolumn{5}{c}{\textbf{Models with External Data/Knowledge}} \vspace{0.2em}\\
% MSMN \citep{yuan-etal-2022-code} & 92.50 &94.39 &67.64 &71.78 &67.23\\
% KEPTLongformer \citep{Yang2022KnowledgeIP} &92.63 &94.76 &68.91 &72.85& 67.26\\
% \bottomrule
% \end{tabular}
% }
% \caption{Results on the MIMIC-III-50 test set. The results are the mean over five runs with different random seeds.}\label{tab:result_common_50}
% \end{table*}
%\vspace{-0.5em}
\subsection{Experiment Settings}
\label{subsec:experiment-settings}
\paragraph{Setup}
We conduct experiments on the data set MIMIC-III \citep{johnson2016mimic}. Following the previous work Joint LAAT \citep{ijcai2020-461-vu}, we consider two versions of MIMIC-III dataset: \textbf{MIMIC-III-Full} consisting of the complete set of 8,929 codes and  (\textbf{MIMIC-III-50}) consisting the 50 most frequent codes.
% The former consists of 52,722 discharge summaries: 47,719 for training, 1,631 for validation, and 3,372 for testing while the latter consists of 11,317 discharge summaries: 8,067 for training, 1,574 for validation, and 1,730 for testing. 
Similarly to \citet{Yang2022KnowledgeIP}, we use macro and micro AUC and F1, as well as precision@k ($k=8$ for MIMIC-III-Full and $k=5$ for MIMIC-III-50). For both data sets, we train with one single 16GB Tesla P100 GPU. We detail relevant training hyperparameters and the statistics of the data sets in the Appendix.
% We list all detailed training hyper-parameters and the statistic of the datasets in the Appendix. The development set is used to find the best-performing thresholds, and we use it to perform inference on the test set.

We compare our models with recent state-of-the-art work using the results from \citet{Yang2022KnowledgeIP}. Among them, Joint LAAT is most similar to our work because it uses a similar attention mechanism and considers both parent and child labels; therefore, we use it as a comparison in ablation studies. We run our models five times with the same hyper-parameters using different random seeds and report the average scores.

%\vspace{-0.5em}
\subsection{Main Experiment Results}
%\vspace{-0.25em}
\paragraph{MIMIC-III-Full}
From the result shown in Table \ref{tab:result_full}, we see that our model achieves a micro F1 of $58.4\%$, the highest among ``single'' models that do not rely on external data/knowledge. Specifically, our model outperforms Joint LAAT by about $2.5\%$, $0.2\%$, $0.9\%$, $0.3\%$, $0.9\%$ in macro AUC, micro AUC, micro F1, macro F1 and precision@8 respectively. In particular, our model is on par with MSMN \citep{yuan-etal-2022-code} which uses code synonyms collected from \citet{umls2004} to improve label embeddings. Moreover, our model is computationally more effective than MSMN with $1.25$ vs $11$ hours per training epoch and $47$ vs. $99$ seconds to infer the dev set on a single P100 GPU. 

Improvements of other models \citep{huang-etal-2022-plm,Yang2022KnowledgeIP} most likely stem from the use of external information in form of knowledge injected into pre-trained language modeling. We leave the integration of such information into our proposed model architecture for future work.%can be beneficial. % We leave this to future work.
%\vspace{-0.25em}
\paragraph{MIMIC-III-50}
From the results on the right-hand side of Table~\ref{tab:result_full}, our model produces a micro-F1 of $71.83\%$, the highest among single models. Specifically, our model surpasses Joint LAAT \cite{ijcai2020-461-vu} with nearly $1.0\%$, $2.0\%$, $0.4\%$ absolute improvement in micro F1, macro F1, and precision@5, respectively. In particular, the macro-F1 of our model is on par with the much more complex state-of-the-art method KEPTLongformer~\citep{Yang2022KnowledgeIP}. This demonstrates the ability of our model to be adapted to classification problems with a large or small number of labels while having competitive results in both cases.
%\vspace{-0.5em}
\subsection{Ablation Study}
\begin{figure}[!t]
    \begin{tikzpicture}
    \begin{axis}[
        xbar=-10pt,
        y tick label style={anchor=north east, align=right,text width=1cm, font=\scriptsize\itshape},
        legend style={font=\small},
        height=10.5em,
        major y tick style = transparent,
        % ylabel=\emph{Label Frequency},
        % xlabel=\emph{Micro F1},
        width  = \columnwidth,
        bar width=4pt,
        xmajorgrids=true,
        %y tick label as interval,
        %ymajorgrids=true,
        x grid style=dashed,
        legend pos=south east,
        legend cell align={left},
        legend columns=1,
        ytick={1,2,3,4,5},
        ymin=-0.1, ymax=5.1,
        xmax=75,
        nodes near coords,
        every node near coord/.append style={font=\tiny},
        yticklabels={1-10, 11-50, 51-100, 101-500, >500},
    ]
    \addplot[legend entry=\textsc{Ours}, color=bblue, fill=bblue]  coordinates {(4.2,0.47) (24.2,1.47) (35.9,2.47) (46.9,3.47) (67.1,4.47)};
     \addplot+[legend entry=\textsc{JointLAAT}, color=rred, fill=rred]  coordinates {(3.4,0.53) (23.5,1.53) (34.9, 2.53) (45.9,3.53) (66.5,4.53)};
    \end{axis}
    \end{tikzpicture}
    %\vspace{2\baselineskip}
    
    \caption{Comparison of Micro-F1 scores between our model and JointLAAT on labels with different Mimic-III-Full testset frequencies.}
    \label{fig:Micro F1 by label frequency group}
\end{figure}
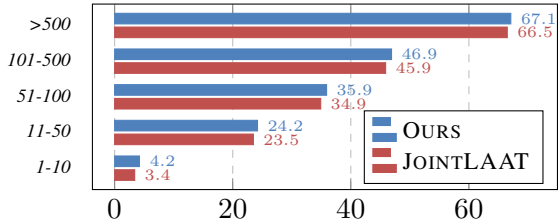
% (Similar to https://aclanthology.org/2022.findings-acl.110.pdf)
To evaluate the effectiveness of our model, we conduct an ablation study on the MIMIC-III-Full set, comparing it with Joint LAAT. Rather than integrating parent label prediction scores as supplementary features with the child label representation, as done in the Joint LAAT method, we allow child label representations to attend to both parent label and document representations. We show that this approach drives performance improvements in two aspects: parent label prediction and performance on labels grouped by frequency of appearance.
%\vspace{-0.5em}
\paragraph{Parent Label Prediction} Table \ref{tab: Parent Label Prediction} compares the results of parent label prediction. Our model outperforms Joint LAAT by $0.9\%$, $0.9\%$, and $0.7\%$ absolute in macro F1, micro F1, and Precision@8, which naturally yields in better child label prediction performance reported in previous sections. %Obviously, when we predict the parent codes more correctly, we predict the child codes better. 
But even considering only the case where both models predict parent labels correctly, our approach still achieves a micro F1 score of $65.5\%$,  outperforming Joint LAAT with a micro F1 score of $65.0\%$. This demonstrates that both parent code and child code prediction benefit from our approach. 
% We include examples where our model predicts the same parent labels as Joint LAAT but predicts child labels better in the Appendix.
% \begin{table}
% \centering
% \resizebox{0.7\columnwidth}{!}{%
% \begin{tabular}{lrr}\toprule
% Metrics & Our model & Joint LAAT \\
% \midrule
% Macro F1 &29.1 & 28.2\\
% Micro F1 &69.0 & 68.1\\
% Micro AUC & 98.7 & 97.8\\
% Macro AUC & 93.4 & 92.3\\
% P@8 & 83.0 & 82.3\\
% \bottomrule
% \end{tabular}
% }
% \caption{Parent prediction results in MIMIC-III-Full.}\label{tab: Parent Label Prediction}
% \end{table}
\begin{table}[!t]
\centering
\resizebox{1\columnwidth}{!}{%
\begin{tabular}{lrrrrr}
%\toprule
\multirow{2}{*}{} &\multicolumn{2}{c}{AUC} &\multicolumn{2}{c}{F1} & Precision \\
\cmidrule{2-6} &Macro &Micro &Macro &Micro &P@8 \\\midrule
Ours & 98.7 & 93.4 & 29.1 & 69.0 & 83.0 \\
JointLAAT & 97.8 & 92.3 & 28.2 & 68.1 & 82.3 \\
\bottomrule
\end{tabular}
}
\caption{Parent prediction results on MIMIC-III-Full.}\label{tab: Parent Label Prediction}
\end{table}
\vspace{-0.25em}
\paragraph{Performance in Label Frequency Groups}

To understand more about our prediction of the model, we divide medical codes into five groups based on their frequencies in MIMIC-III-Full: $1-10, 11-50, 51-100, 101-500, >500$ like \citet{wang-etal-2022-novel}. We list the statistics of all groups in the Appendix. We compare the micro F1 between different groups in Figure~\ref{fig:Micro F1 by label frequency group}. Overall, we outperform Joint LAAT in all groups. The relative improvements are most noticeable in the rare-frequency group ($25\%$ relative improvement in the $1-10$ group, vs $2\%$ or less in other cases). A possible explanation for this is that the parent label space is smaller than the full label space, which results in more training samples per parent label, allowing to learn better representations. As the parent label representation is used to compute child label representations, low-frequency child labels can thus benefit from representations learned from their high-frequency siblings. %It shows that our model can cope with the rare code groups well.
\section{Conclusion}
%\vspace{-0.5em}
In this paper, we have presented a novel, simple but effective two-stage decoding model that leverages the hierarchical structure of the ICD codes to decode from parent-level codes to child-level codes. Experiments on the MIMIC-III data set show that our model outperforms other single-model work and achieves on-par results with models using external data/knowledge. Our ablation studies validate the effectiveness of our model in predicting the code hierarchy and codes in different frequency groups. In future work, we intend to integrate our decoder with a better document or label representation to further improve performance.
% Entries for the entire Anthology, followed by custom entries

\clearpage
\section*{Limitations}

As established, medical coding is an important task for the healthcare industry. Efforts toward its successful automation have wide-ranging implications, from increasing the speed and efficiency of clinical coders while reducing their errors, saving expenses for hospitals, and ultimately improving the quality of care for patients.

However, with this goal in mind, our study presented here should be contextualized by the two main limitations that we identify and outline below.

As with other data-driven approaches, the evaluation performance discussed in our paper is limited by the choice of the (static) MIMIC-III data set. This data set could be seen as lacking diversity, as it only features a fraction of all possible ICD-9 codes and contains medical notes collected in English from a specific department of patients belonging to a specific demographic.
While our approach does not make any explicit assumptions about the nature of the data other than the hierarchy of labels, in absence of formal guarantees, we cannot make rigorous statements about the efficacy of our (or indeed any related) approaches on clinical data gathered in different settings, such as other languages, countries or departments.

The second limitation is of a more practical nature, since 2015 the ICD-9 coding system is being phased out in favor of the more expressive ICD-10 system, thus ICD-9 coding has limited applications in practice. However, as with its predecessor, the ICD-10 codes are organized in an even richer hierarchy, which should enable the successful application of our proposed approach. 

\bibliography{anthology,custom}
\bibliographystyle{acl_natbib}

%\newpage
\appendix
\section*{Appendix}
Table~\ref{tab:dataset} depicts the dataset statistic of the MIMIC-III-Full and MIMIC-III-50 datasets, Table~\ref{hyper para} details the choice of hyperparameters of our best-performing model and Table~\ref{tab: label frequency distribution} outlines the characteristics of the label frequency groupings for the Ablation study in Section~3.3.
%\section*{Dataset Splits and Details}
\begin{table}[ht]
    \centering
    \small
    \begin{tabular}{lccc}
    \toprule
    & Train & Dev & Test  \\
    \multicolumn{4}{c}{MIMIC-III Full} \\
    \midrule
    \# Doc. & 47,723 & 1,631 & 3,372 \\
    Avg \# words per Doc. & 1,434 & 1,724 & 1,731 \\
    Avg \# parent codes per Doc. & 13.7 & 15.4 & 15.9 \\ 
    Total \# parent codes & 1149 & 741 & 850 \\
    Avg \# child codes per Doc. & 15.7 & 18.0 & 17.4 \\ 
    Total \# child codes & 8,692 & 3,012 & 4,085\\ 
    \midrule
    \multicolumn{4}{c}{MIMIC-III 50} \\
    \midrule
    \# Doc. & 8,066 & 1,573 & 1,729 \\
    Avg \# words per Doc. & 1,478 & 1,739 & 1,763 \\
    Avg \# parent codes per Doc. & 5.3 & 5.6 & 5.7 \\ 
    Total \# parent codes & 39 & 39 & 39 \\
    Avg \# child codes per Doc. & 5.7 & 5.9 & 6.0 \\ 
    Total \# child codes & 50 & 50 & 50 \\
    \bottomrule
    
    \end{tabular}
    \caption{Statistics of MIMIC-III dataset under full codes and top-50 codes settings.}
    \label{tab:dataset}
\end{table}
%\section*{Hyper-parameters Details}
\begin{table}[ht]
\centering
\small
\begin{tabular}{lc}
\toprule
Parameters & Value \\
\midrule
Emb. dim. ($d_e $) & 100 \\
% Emb. dropout & 0.2 & 0.2 \\
LSTM Layer & 1\\
LSTM hidden dim. ($h$)& 512\\
LSTM output dim. ($d$) & 512\\
% Synonyms count ($M$) & 4 & 8 \\
% Rep. dropout & 0.3 & 0.2 \\
% R-Drop weight & 5.0 & 5.0 \\
Epoch & 50\\
% Peak lr. & 0.001 & 5e-4 \\
Batch size & 8\\
% Adam $\epsilon$ & 1e-8 & 1e-8 \\
% Weight decay & 0.01 & 0.01 \\
% Clipping grad. & 1.0 & 1.0 \\
Maximum Sequence Length & 4000\\
Optimizer & Adam\\
\bottomrule
\end{tabular}
\caption{Hyper-parameters used for training MIMIC-III full and MIMIC-III 50.
}
\label{hyper para}
\end{table}

%\section*{Label Frequency Group Details}
\begin{table}[ht]
\centering
\small
\begin{tabular}{rrr}\toprule
\multirow{2}{5em}{Frequency range} & \multirow{2}{5em}{Number of parent codes} & \multirow{2}{5em}{Number of child codes} \\
& & \\
\midrule
1-10 &301 & 5394\\
11-50 &270 & 1872\\
51-100 & 101 & 549\\
101-500 & 246 & 797\\
>500 & 240 & 309\\

\bottomrule
\end{tabular}
\caption{Label Frequency Distribution}\label{tab: label frequency distribution}
\end{table}

\end{document}

%% file: math_commands.tex
\usepackage{amsmath}
\usepackage{amsfonts,bm}
\makeatletter   
\newcommand{\sveryshortarrow}[1][3pt]{\mathrel{%
    \vcenter{\hbox{\rule[-.5\fontdimen8\scriptfont3]
               {\scriptratio\dimexpr#1\relax}{\fontdimen8\scriptfont3}}}%
   \mkern-4mu\hbox{\let\f@size\sf@size\usefont{U}{lasy}{m}{n}\symbol{41}}}}
\makeatother

\def\eqref#1{equation~\ref{#1}}
\def\1{\bm{1}}

\def\vx{{\bm{x}}}

\def\m1{{\bm{1}}}

\def\mH{{\bm{H}}}

\def\mL{{\bm{L}}}

\def\mP{{\bm{P}}}

\def\mV{{\bm{V}}}
\def\mW{{\bm{W}}}

\DeclareMathAlphabet{\mathsfit}{\encodingdefault}{\sfdefault}{m}{sl}
\SetMathAlphabet{\mathsfit}{bold}{\encodingdefault}{\sfdefault}{bx}{n}

\def\sL{{\mathbb{L}}}

\def\sP{{\mathbb{P}}}

\def\sR{{\mathbb{R}}}

\newcommand{\Ls}{\mathcal{L}}
\newcommand{\R}{\mathbb{R}}

\newcommand{\softmax}{\mathcal{S}}

\newcommand{\sigmoid}{\sigma}

%% file: macro.tex
\usepackage{xspace}